%% file: ms.tex
\documentclass[letterpaper,10 pt,conference]{ieeeconf}
\IEEEoverridecommandlockouts
\usepackage{cite}
\usepackage[top=57pt,left=48pt,right=48pt,bottom=46pt]{geometry}
\usepackage{amsmath,amssymb,amsfonts}
\usepackage{algorithmic}
\usepackage{graphicx}
\usepackage{textcomp}
\usepackage{tabularx}
\usepackage{xcolor}
\usepackage{dsfont} %kre
\usepackage[colorlinks=true, linkcolor=black, urlcolor=black]{hyperref} %kre

\input{customOperatorsAndVariables.tex} %kre
\def\BibTeX{{\rm B\kern-.05em{\sc i\kern-.025em b}\kern-.08em
    T\kern-.1667em\lower.7ex\hbox{E}\kern-.125emX}}
\begin{document}

\title{\LARGE \bf Enhancing Voluntary Motion with Modular, Backdrivable, Powered \\Hip and Knee Orthoses\\
\thanks{This work was supported by the National Science Foundation under Award Numbers 1949869 and 1953908, and by the National Institute of Biomedical Imaging and Bioengineering of the NIH under Award Number R01EB031166. The content is solely the responsibility
of the authors and does not necessarily represent the official views of the NIH.}
\thanks{C. Nesler, G. Thomas, N. Divekar, and R. D. Gregg are with the Department of Electrical Engineering and Computer Science and the Robotics Institute, University of Michigan, Ann Arbor, MI 48109. E. J. Rouse is with the Department of Mechanical Engineering and the Robotics Institute, University of Michigan, Ann Arbor, MI 48109. Contact: {\tt\footnotesize \{neslerc,gcthomas,ejrouse,rdgregg\}@umich.edu}}%
}

\author{Christopher Nesler, Gray Thomas, Nikhil Divekar, Elliott J. Rouse, and Robert D. Gregg}

\maketitle

\begin{abstract}
Mobility disabilities are prominent in society with wide-ranging detriments to affected individuals. Addressing the specific deficits of individuals within this heterogeneous population requires modular, partial-assist, lower-limb exoskeletons. This paper introduces the Modular Backdrivable Lower-limb Unloading Exoskeleton (M-BLUE), which implements high torque, low mechanical impedance actuators on commercial orthoses with sheet metal modifications to produce a variety of hip- and/or knee-assisting configurations. Benchtop system identification verifies the desirable backdrive properties of the actuator, and allows for torque prediction within $\pm$0.4~Nm. An able-bodied human subject experiment demonstrates that three unilateral configurations of M-BLUE (hip only, knee only, and hip-knee) with a simple gravity compensation controller can reduce muscle EMG readings in a lifting and lowering task relative to the bare condition. Reductions in mean muscular effort and peak muscle activation were seen across the primary squat musculature (excluding biceps femoris), demonstrating the potential to reduce fatigue leading to poor lifting posture. These promising results motivate applications of M-BLUE to additional subject populations such as hip/knee osteoarthritis and geriatric frailty, and the expansion of M-BLUE to bilateral and ankle configurations.

\end{abstract}

\section{Introduction}

One in eight adults in the U.S. has a mobility impairment that limits their social activity, economic productivity, and quality of life \cite{CDC:prevalence}. These deficits often stem from advanced age, stroke, or musculoskeletal disorders. Most of these individuals have some voluntary control of their lower limbs, and thus require only \textit{partial} assistance to overcome these deficits. However, no broadly applicable intervention fits this important need \cite{grimmer2019mobility}. Conventional orthoses tend to immobilize rather than assist joints, causing side-effects that include compensatory movements, gait asymmetry, and overdependence on existing orthoses. In addition, most mobility aids (\textit{e.g.,} passive orthoses, canes, and walkers) cannot provide net-positive mechanical work, which presents an ever greater challenge for assisting activities that require substantial energy, including squatting, sit-to-stand, and stair ascent. 

Emerging powered orthoses (\textit{i.e.}, exoskeletons) have the potential to restore normative leg biomechanics \cite{grimmer2019mobility}, but these devices have not been widely adopted in part because of their stiffness, bulkiness, and population-specific designs. Commercial rehabilitation exoskeletons like those manufactured by Ekso Bionics, Indego, and ReWalk feature an integrated lower-body frame with a fixed set of actuators targeting specific joints for a specific patient population \cite{baunsgaard2018gait,goldfarb:exo,zeilig2012safety}. These devices are generally designed with actuators that prevent the user’s weight from backdriving the joint and enable large output torques to provide the complete limb function for a person with spinal cord injury \cite{zeilig2012safety} or severe stroke \cite{Murray2015}. To minimize size and mass, these actuators are designed with small, low-torque, high-speed motors that require a large gear ratio (typically greater than 50:1 \cite{Zhu2021}) to achieve human-scale torques-velocity regimes. The motor’s inertia reflected through the transmission scales with the gear ratio squared, and friction from meshing parts gets similarly amplified \cite{seok2014design}. This results in high mechanical impedance at the joint, which means the user cannot easily backdrive the actuator to move their joints. In this actuation paradigm, the user is typically forced to follow the robot's pre-defined joint patterns rather than their own volition \cite{Tucker2015}.% High-ratio transmissions also have more/faster meshing parts, causing acoustic noise that makes robotic devices less desirable for everyday use.

Recent work has focused on augmenting voluntary motion with backdrivable powered orthoses \cite{Mooney2014,Lv2018,Wang2018,yu2020quasi,Zhu2021} and soft exosuits \cite{asbeck2015multi,Kim2019,awad2017soft,yu2019design}, achieving backdrivability in different ways. Load cells \cite{asbeck2015multi} or series elastic elements \cite{shepherd2017design} can provide force feedback to control torque (as in B-Temia Keeogo \cite{maclean2019energetics}), but this tends to increase mechanical complexity and limit actuator torque bandwidth. The Honda \cite{kusuda2009quest} and Samsung \cite{lee2017gait} hip orthoses use a lower gear ratio to increase backdrivability at the cost of output torque (less than 10 Nm), motivating the use of emerging high-torque, backdrivable actuators called quasi-direct drive actuators. Quasi-direct drive actuators use high-torque ``pancake'' motors to achieve high output torques with low gear ratios (and thus low friction/inertia). These actuators have enabled legged robots to perform dynamic walking, running, and jumping with high control bandwidth, precise current-based torque control, energy regeneration, compliance to impacts, and reduced acoustic noise \cite{seok2014design,kenneally2016design,elery2020a}. We recently implemented quasi-direct drive actuators in leg orthoses \cite{Lv2018,Zhu2021}, enabling the use of an emerging task-invariant control method called \emph{energy shaping} \cite{Lin2021}, \textit{e.g.}, compensating body weight \cite{Divekar2020} and/or inertia \cite{lv2021trajectory}. However, these backdrivable designs do not offer a reconfigurable solution for different use cases (\textit{i.e.}, different joints).

\begin{figure}[t]
\centering
\includegraphics[trim=1 1 1 1, clip, width = 0.95\columnwidth]{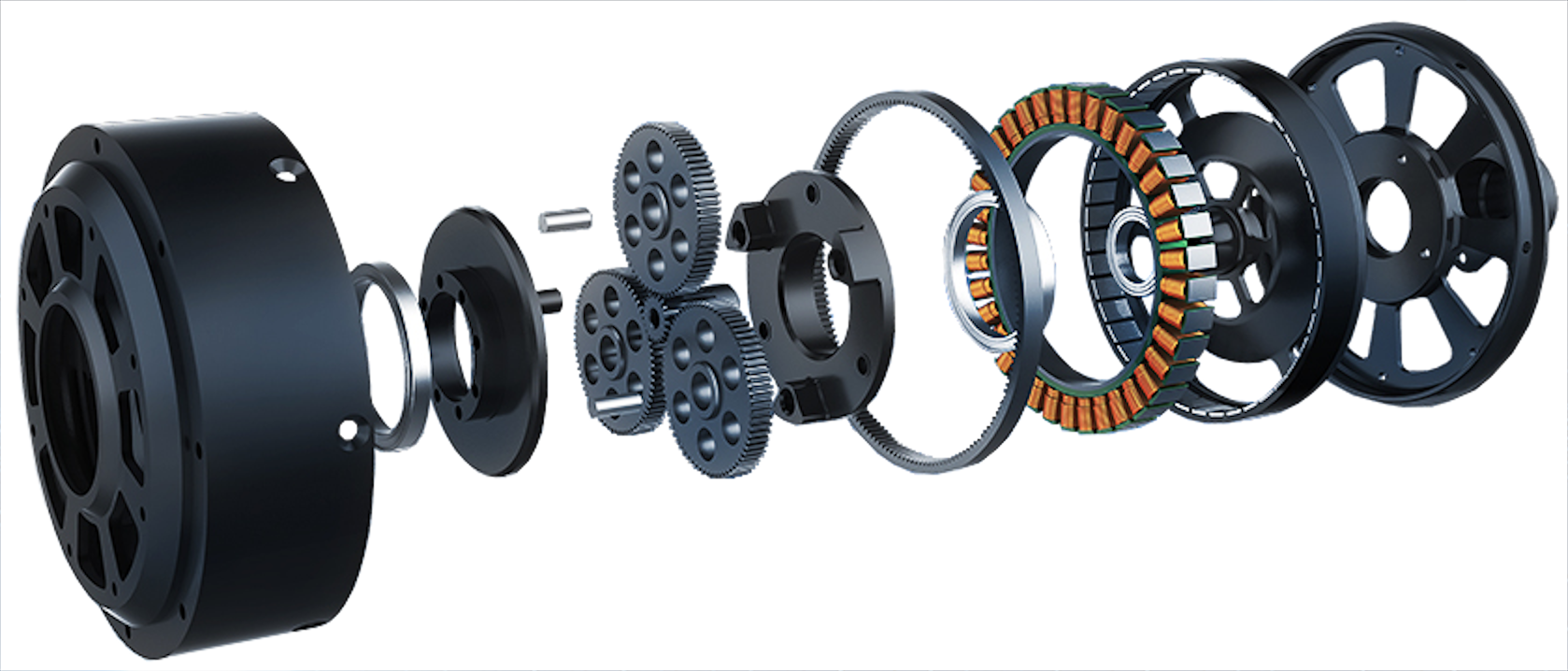}
\vspace{-2mm}
\caption{Exploded view of T-motor A80-9 actuator, an open-frame version of the AK80-9 \cite{AK809}.} \label{fig:AK809}
\vspace{-2mm}
\end{figure}

One of the first modular exoskeletons \cite{brackx2014design} adds a compliant actuator to conventional hip, knee, and ankle orthoses. Although this design allows different joint configurations, the use of series elasticity increases size and mass (1.4 kg per actuator) and limits backdrivability. Additionally, its peak torque of 16.7 Nm may be insufficient for some use cases, such as compensating for the difference in knee torques between a safe squat lift and a risky stoop lift (around 30 Nm \cite{hwang2009lower}) to prevent workplace injuries. Instead of modifying conventional orthoses, custom exoskeleton modules have been designed in \cite{meijneke2021symbitron} to provide substantial torque at the ankle (102 Nm), knee (69 Nm), and/or hip (102 Nm) for individuals with spinal cord injury. Although this system offers a trajectory-free neuromuscular control strategy, its backdrivability is also limited by the use of series elasticity with large gear ratios (up to 100:1). The modular exoskeleton in \cite{li2020design} can achieve similar torques at the knee and hip to assist spinal cord injury, using a mix of custom and conventional orthotic components. This design does not have active torque sensing to compensate for its high gear ratio (up to 120:1), reducing torque control accuracy and backdrivability.

These prior works motivate our use of more backdrivable, compact quasi-direct drive actuators in the presented hip and/or knee system called the Modular Backdrivable Lower-limb Unloading Exoskeleton (M-BLUE). M-BLUE is designed to provide sufficient output torque to partially assist mild to moderate impairments, while having negligible backdrive torque to facilitate voluntary motion. We use the recently released T-Motor AK80-9 actuator (with a 9:1 internal transmission \cite{AK809}) to improve upon our prior design \cite{Zhu2021} with a single-stage planetary gearset inside a torque-dense ``pancake'' Brushless DC motor, enabling high output torque and low backdrive in a compact package. We extend this actuation paradigm to the modular design philosophy of \cite{brackx2014design}, making simple aftermarket modifications to conventional hip and knee orthoses that can be replicated by clinicians. We focus on the hip and knee because conventional ankle-foot orthoses tend to immobilize the ankle, requiring custom orthosis designs like the Dephy ExoBoot \cite{Mooney2014,Ingraham2020}. We also implement a modular, trajectory-free, gravity-compensation control strategy to assist volitional motion at the hip and/or knee. Benchtop experiments demonstrate the lightweight actuator module (584 g) can produce up to 30 Nm with less than 2 Nm of dynamic backdrive torque during human-like walking motions. We demonstrate that the hip-only, knee-only, and hip-and-knee joint configurations reduce muscle activity during an able-bodied lifting and lowering experiment, suggesting the presented system can reduce muscular fatigue leading to risky lifting posture \cite{trafimow1993effects}. By leveraging the Dephy FASTER embedded system, the hip and knee modules can be easily interfaced with the Dephy ExoBoot system to enable ankle configurations in future work.

\section{Design and Control} \label{Sec:Design}
\begin{figure*}[t]\centering
\resizebox{.8\textwidth}{!}{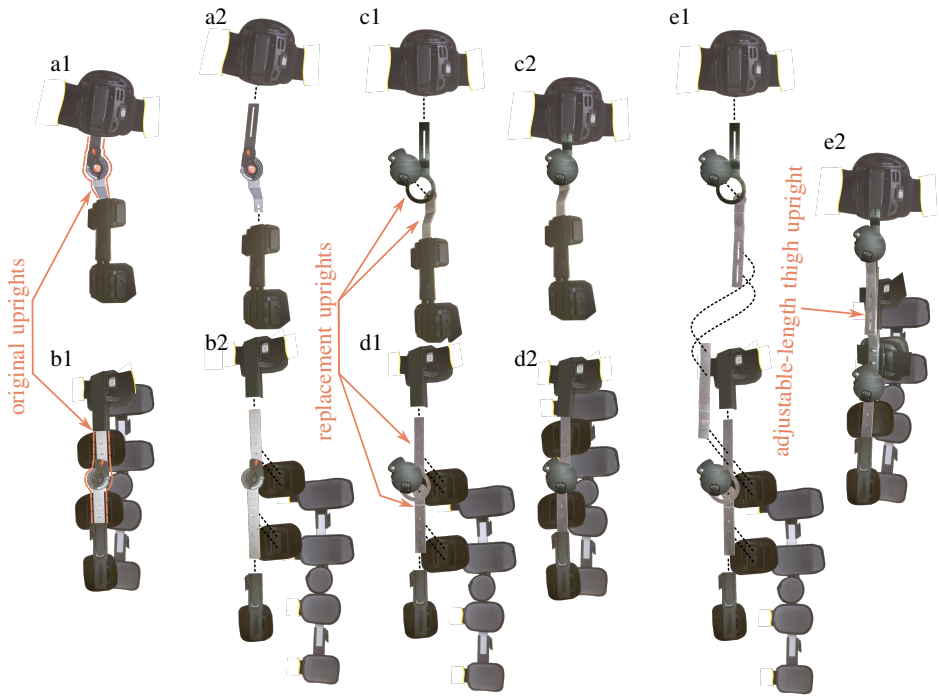}
\caption{The Modular Backdrivable Lower-limb Unloading Exoskeleton (M-BLUE): a1-2) T Scope Hip brace disassembly, b1-1) T Scope Knee brace disassembly, c1-2) M-BLUE hip-only assembly, d1-2) M-BLUE knee-only assembly, and e1-2) M-BLUE hip-knee assembly.\label{fig:MBLUE}}
\end{figure*}
\subsection{Actuator Selection}

We selected the AK80-9 actuator (T-motor, China) for its high torque density, low gear ratio (9:1), low weight (485 g), and low cost (\$580 USD). The actuator assembly sets a 9:1 planetary gear inside a wide diameter BLDC motor similar to those used in high-power electric quadcopters (see Fig.~\ref{fig:AK809},  \cite{AK809}).
Our previous study on the thermal properties and performance of the T-motor U8-KV100, which is nearly identical to the electric motor inside the AK80-9 assembly, demonstrated its excellent torque/speed capabilities for legged robot applications \cite{lee2019empirical}.
In particular, the analysis suggests that the actuator can provide torque up to 30 Nm, which is 25-50\% of normative peak torque depending on the joint, task, and subject weight \cite{winter2009biomechanics}. 
This torque output can 1) compensate for the difference in knee torque required for a safe squat lift (``lift with the knees”) compared to a riskier stoop lift posture (``lift with the back”) \cite{hwang2009lower}, 2) overcome the difference in knee moments between the affected and unaffected sides of an average individual with chronic stroke during sit to stand \cite{roy2007side}, and 3) reach the strength threshold needed to assist ageing individuals with activities of daily living (ADLs) \cite{grimmer2019mobility}. 

\subsection{Modular Design}

% Commercial orthoses were the basis of the design
In keeping with the modular, clinically-approachable vision for this exoskeleton, commercially available orthoses were used as the basis of the mechanical human-robot interface. We used the T Scope Premier Post-Op Knee Brace (see Fig.~\ref{fig:MBLUE}.b) and T Scope Post-Op Hip Brace (see Fig.~\ref{fig:MBLUE}.a) (Breg, California, USA) to construct a modular system (Fig.~\ref{fig:configs}) that combines our own joints and custom upright structure with the industrially designed human-attachment interfaces, adjustment features, and straps of the medical orthoses. We used readily available tools (\textit{e.g.}, drill press, circular punch, Dremel, chisel) to partially disassemble the orthoses and replace key structural components. This will make future replication of these modifications a straightforward process for clinicians interested in implementing this modular powered orthosis system.

% The upright components of the braces were replaced with custom hardware
The two braces share a similar structure in which lateral metal uprights hinge near the anatomical joint axis, and interfacing pads are attached to these uprights with rivets or custom molded plastic features. The knee brace has a second set of uprights on the medial side of the leg. We converted the orthoses by first harvesting the pads connected to the lateral uprights. These four metal upright components (two from the knee brace, two from the hip brace) were replaced by analogous custom parts that interface with the input and output structures of the actuators (see Fig.~\ref{fig:MBLUE}). We retained the original medial uprights from the knee brace for use in the knee-only and hip-knee configuration (Fig.~\ref{fig:MBLUE}.d,e). The distal upright for the hip joint has two configurations: one for connecting with a knee brace (Fig.~\ref{fig:MBLUE}.e), and one for independent hip use (Fig.~\ref{fig:MBLUE}.c). The knee-brace configuration allows for length adjustment of a thigh upright that attaches to the proximal knee upright (maintaining the pad configuration and adjustment mechanism of the proximal knee brace), while the hip-only configuration retains the original hip pad and adjustment mechanism of the distal hip brace. 

% the parts are sheet metal
These custom parts (Fig.~\ref{fig:anglesCAD}) were fabricated from 1/8" sheet aluminum (7075-T6, chosen for its high strength-to-weight ratio) primarily using a CNC router and a sheet metal brake for bending. While the tools needed to manufacture sheet metal parts such as these are not part of a traditional clinical setup, they are easy to manufacture in commercial shops and could be distributed in kits similar to the way many of the metal components in a traditional custom orthosis can be ordered from a catalog. The actuators are fastened to the sheet metal parts (Fig.~\ref{fig:anglesCAD}) using bolts and pins. The riveted connections between pads and uprights are replaced by threaded fasteners, washers and nuts to facilitate disassembly and reconfiguration. The adjustable length connection between the hip and knee structures in the hip-knee configuration (Fig.~\ref{fig:MBLUE}.e) is achieved with shoulder bolts.

% While many engineering schools and other mechanically-oriented work environments may provide access to these resources, we acknowledge that CNC routers and sheet metal brakes are likely outside the scope of equipment available to orthotists who may eventually want to implement a powered orthosis in this manner. By constraining the design of custom-fabricated metal parts to rely solely on basic sheet metal fabrication techniques, we anticipate the additional cost of out-of-house manufacturing will remain reasonable for teams without direct access to CNC routers or sheet metal brakes.

% Pieces are assembled with screw-based fastening
The design emphasizes modularity and the ability to convert between the various configurations easily. The knee-only configuration (Fig.~\ref{fig:MBLUE}.d) utilizes a subset of the sheet-metal components from the hip-knee configuration (Fig.~\ref{fig:MBLUE}.e) along with their affiliated fasteners (nuts, bolts, pins, and washers).  The hip-only configuration (Fig.~\ref{fig:MBLUE}.c, center) uses a different set of distal components from the hip-knee configuration (distal components from the hip brace in Fig.~\ref{fig:MBLUE}.a rather than proximal components from the knee brace in Fig.~\ref{fig:MBLUE}.b). Converting from hip-only to hip knee requires replacing only one two-pin 6-screw actuator interface, and the four screws from the adjustable-length thigh upright system.

\subsection{Embedded System and Control}

\begin{figure}
\centering
\resizebox{.9\columnwidth}{!}{
\includegraphics[height = 0.3\textwidth,trim={0 0 10cm 0},clip]{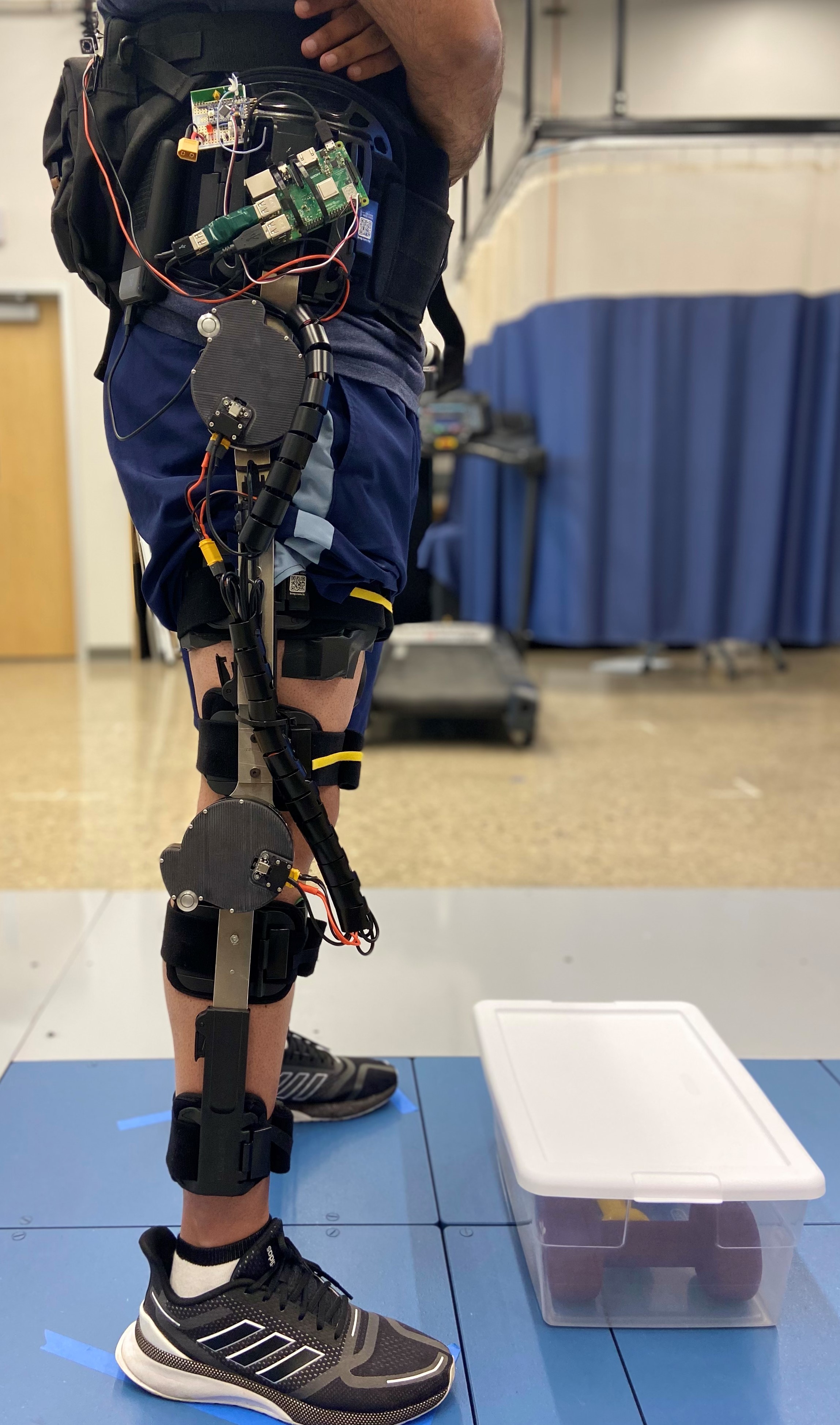}\hspace{2mm}\includegraphics[height = 0.3\textwidth,trim={0 0 10cm 0},clip]{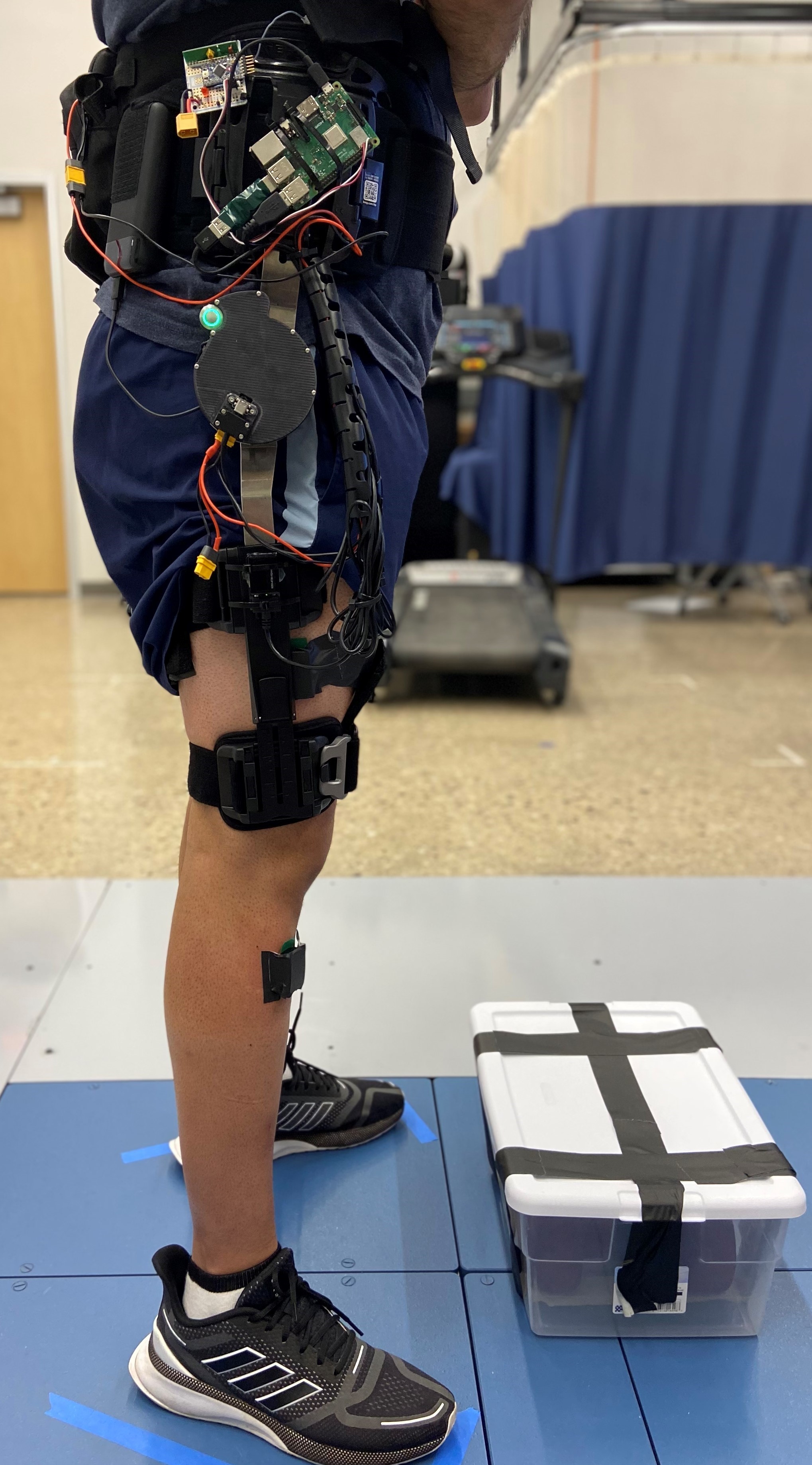}\hspace{2mm}\includegraphics[height = 0.3\textwidth,trim={0 0 10cm 0},clip]{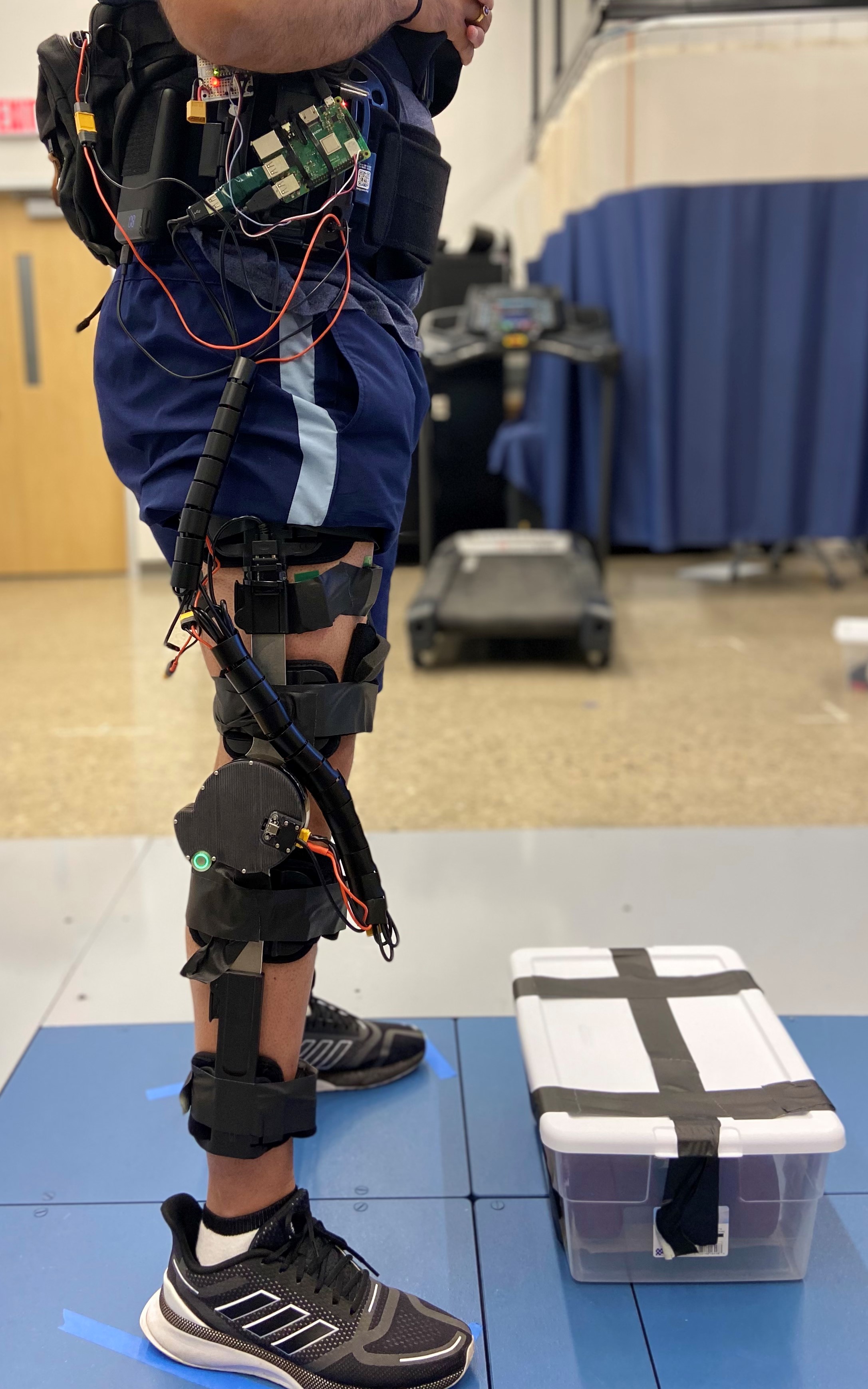}}
\vspace{-1mm}
\caption{Hip-knee, hip only, and knee only configurations of modular powered orthoses during lifting and lowering experiment. Note that the EMG electrodes and top PCB (for wireless sync) are only for assessment purposes, and would be removed during normal use. The Raspberry Pi can also be enclosed in a case for greater robustness.} \label{fig:configs}
\vspace{-2mm}
\end{figure}

% We use FASTER
The AK80-9 actuators were combined with the FASTER motor driver system by Dephy, Inc. (Maynard, MA), for a total actuator weight of 584 g. Based on \cite{Duval2016, Duval2016a}, the FASTER driver system features a high-bandwidth current control mode (corresponding to torque) and communicates via USB to a central Raspberry Pi 3B+ embedded computer (Raspberry Pi Foundation, Cambridge, United Kingdom) in a 300~Hz real-time loop. The Raspberry Pi and battery are mounted at the waist. The associated software application programming interface allows us to control the actuator in Python. The electrical system uses two lithium polymer batteries mounted in a hip-bag (Fig.~\ref{fig:configs}).

\begin{figure}
\centering
\includegraphics[trim=100 000 100 0, clip, width = 0.9\columnwidth]{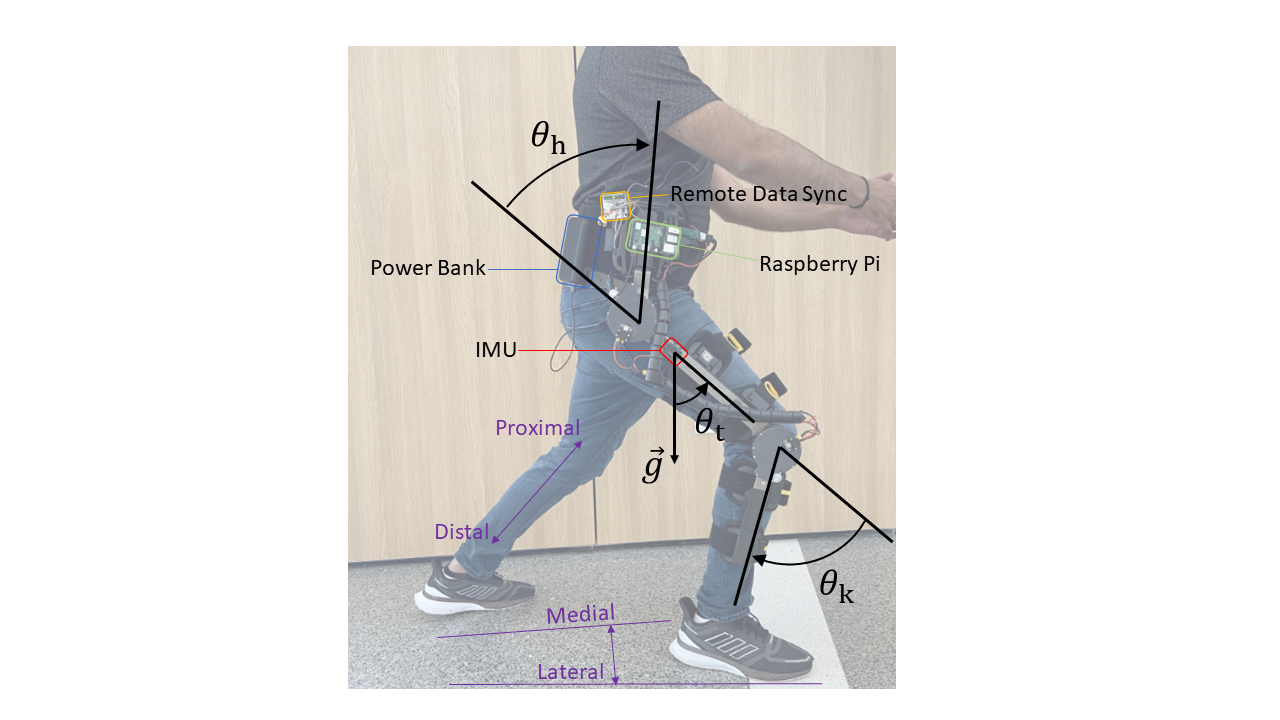}
\vspace{-2mm}
\caption{M-BLUE hip-knee configuration with superimposed joint angle conventions for hip ($\theta_{h}$) and knee ($\theta_{k}$) angles, provided by actuator encoders, as well as thigh angle ($\theta_{t}$) with respect to gravity (g), provided by the IMU (red). Other annotated systems include the Raspberry Pi (green), its power source (blue), and a remote data sync board (orange; only required during EMG experiments). A convention for directional vocabulary (purple) is also included for reference.} \label{fig:anglesCAD}
%\vspace{-1mm}
\end{figure}

% We sense angles and compensate gravity
The controller has access to absolute motor encoders on the actuator models, as well as an inertial measurement system (3DM-GX5-25, LORD Microstrain, Vermont, USA) that provides a global rotation matrix for the the orientation of the thigh segment of the orthosis. Using the assumption that the z-axis of the IMU is perpendicular to the sagittal plane, we convert this rotation matrix to a global sagittal-plane angle for the thigh, and---using the hip encoder---a measurement of the global angle of the waist belt/torso. Thus, we provide the controller with the hip angle $\theta_{h}$, knee angle $\theta_{k}$, and global thigh angle $\theta_t$ data as defined in Fig.~\ref{fig:anglesCAD}.

% We apply a simple gravity shaping controller
To demonstrate the assistive capabilities of M-BLUE, we implemented a simple potential-energy shaping controller \cite{Divekar2020} based on a sagittal plane model of the human dynamics. This controller offloads the gravitational torques from a point mass  which represents a fixed percentage $\alpha$ of the wearer's recorded weight $m$ using the M-BLUE actuators. This controller is based on a modular, stance-only, 1- or 2-link kinematic chain model (depending on the M-BLUE configuration) starting from the most distal link connected to an actuator and working up. This means that for configurations with a powered knee, the shank is considered the base link, and in the hip-only configuration, the base joint is the thigh link. The point mass is located in the most proximal link connected to an actuator: in the torso link for the hip and hip-knee configuration and in the thigh link for the knee-only configuration. 

% The control law.
The control laws for the different configurations result directly from trigonometry. For example, the control law in the hip-knee configuration is
\begin{align}
    \tau_k & = m g \alpha (l_t \sin(\theta_t)-l_h \sin(\theta_t-\theta_h)),\\
    \tau_h & = m g \alpha (-l_h \sin(\theta_t-\theta_h)),
\end{align}
where $g=9.81~\mathrm{m}/\mathrm{s}^{2}$ represents gravitational acceleration, $l_t$ represents the length of the thigh link, and $l_h$ represents the height of the torso-fixed point mass above the hip center. We use the standing position to calibrate zero angles for all of $\theta_h$, $\theta_k$, and $\theta_t$, so we can guarantee zero torques in the standing configuration. The locations for the offloaded point mass were tuned by hand to assist during squat lifting: $l_h$ = 177.8 mm into the torso for configurations with a hip joint, or at the hip center for the knee-only configuration (implemented as $l_h=0,\ \theta_h=0$). The thigh length was set to a default value of $l_t$ = 457.2~mm, but can be easily adjusted. Due to the nature of the control law, the controller is robust to discrepancies between the $l_t$ input magnitude and anatomical value; it can be seen as a slight modification of the assistance ratio $\alpha$. For safety, the actuator torques were saturated at 25 Nm in extension, and at 0 Nm in flexion.

% The control law uses the hip$\theta_{h}$, knee angle $\theta_{k}$, and global thigh angle $\theta_t$ as defined in Fig.~\ref{fig:anglesCAD}, found via measurements from the IMU and built-in actuator encoders, along with gravity (g=9.81~m/s$^{2}$), subject mass $m$, mass-fraction $\alpha$, and hand-tuned values for the length between the subject's hip and center of mass, {$l_{h}$}, and thigh length (between hip and knee joints), \mathrm($l_{t}$) .
% The subject's mass was modeled as a point mass located seven inches (NOTE check with Gray) above the hip joint.

% \begin{equation} \label{hipEQ}
%     \tau_{knee} = mg(l_{t} sin(\theta_{t}) + l_{h} sin(\theta_{h}))
% \end{equation}

% \begin{equation} \label{kneeEQ}
%     \tau_{hip} = mg(l_{h} sin(\theta_{h}))
% \end{equation}

\section{Benchtop Actuator Testing} \label{sec:Bench}

We assessed the AK80-9 actuator's performance in several capacities pertinent to our application. This included static and dynamic backdrive torque tests to demonstrate the mechanical impedance of the system when it is not actively providing assistive torque, as well as friction characterization to inform our open-loop current control model. Friction characterization data were collected using a custom benchtop dynamometer setup which includes a rotary load cell, misalignment couplings, and an antagonist actuator. The antagonist actuator consisted of a T-Motor U8 with a 50:1 transmission (Boston Gear, Charlotte, North Carolina), also driven using a Dephy FASTER motor driver system. Data were recorded from the actuators and the single-axis torque sensor (TRS605, FUTEK, Irvine, California; 16-bit ADS1115 ADC) using a Raspberry Pi 3B+ embedded computer.

\begin{figure}
\centering
% \includegraphics[width = 0.4\textwidth]{dynBDplots.png}
% \vspace{-2mm}
\resizebox{.99\columnwidth}{!}{\scriptsize\def\svgwidth{.99\columnwidth}
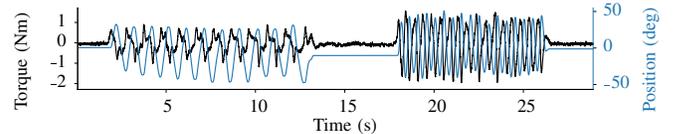}
\vspace{-6mm}
\caption{Torque and position measurements from dynamic backdrive benchtop experiment.} \label{fig:dynBD}
\vspace{-1mm}
\end{figure}

% \begin{figure}%
% \centering%
% \def\svgwith{.9\columnwidth}%
% \resizebox{.95\columnwidth}{!}{\input{DynamicBackdrive.pdf_tex}}
% \vspace{-2mm}
% \caption{Cumulative density histogram from the torque measured in the dynamic backdrive test.} \label{fig:backdrive_histogram}
% %\vspace{-2mm}
% \end{figure}
\subsection{Backdrive Tests}
The static backdrive torque was quantified through manual deflection of the actuator output shaft via analog torque wrench (03727A, Neiko USA, China) from a stationary state until friction was overcome and the output shaft rotated. The static backdrive torque was measured under 0.5~Nm.

The dynamic backdrive torque was investigated by manually deflecting the AK80-9 from the misalignment coupling attached to the opposite side of the load cell (disconnected from the antagonist actuator) at a frequency and amplitude akin to those seen in the knee joint during able-bodied ambulation \cite{winter2009biomechanics}. We selected target trajectories of 1 and 2~Hz with a peak-to-peak amplitude of 70~deg (Fig.~\ref{fig:dynBD}) to approximate the behavior of the knee joint during walking. A 120~beats/min metronome and positional markings on the testbed governed the manual perturbations made for this test. Each phase lasted roughly 10 seconds.

%of ###CYCDEVSEC~Hz with a peak to peak amplitude of approximately ###AMPDEGS~deg (the range of 98\% of histogram values). Over X cycles, a peak backdrive torque of ###BDTQ~Nm was recorded. The deflections in the second phase of the test occurred at an average of ###CYCDEVSEC~Hz with a peak to peak amplitude of approximately ###AMPDEGS~deg (the range of 98\% of histogram values). Over X cycles, a peak backdrive torque of ###BDTQ~Nm was recorded.

The dynamic backdrive torque in this frequency range and amplitude setting was less than 2~Nm (Fig.~\ref{fig:dynBD}). The backdrive torque magnitude increased from the 1~Hz sinusoidal target to the 2~Hz target, consistent with an inertial component of backdrive torque \cite{Zhu2021}.

\begin{figure}[tb]%
\centering%
\rule{0pt}{1mm}\\\resizebox{.99\columnwidth}{!}{\scriptsize\def\svgwith{.99\columnwidth}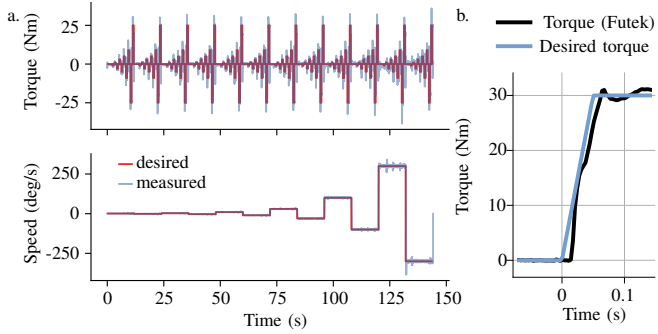}%\includegraphics[width = 0.5\columnwidth]{Benchtest2-NIHpeakTorque.pdf}
\vspace{-2mm}
\caption{Constant torque and speed test (a), and 30~Nm step test (b).} \label{fig:constant_torque_speed}
%\vspace{-2mm}
\end{figure}

\subsection{Constant Torque and Speed Tests}

The actuator was characterized with a non-linear grid of speed and torque values. Speeds included 0, 1, 3, 10, 30, 100, and 300~deg/s, and torques included 0, 1, 3, 5, 9, and 25~Nm. We iterated through all possible combinations of positive and negative speeds and torques from these lists (Fig.~\ref{fig:constant_torque_speed}.a). In all cases, the antagonist actuator controlled speed while the primary actuator controlled current based on the nominal model $\tau\approxeq g_r K_t i$, where $g_r=9$ is the gear ratio, $K_t=0.14$ is the q-axis torque constant \cite{lee2019empirical}, and $i$ is the q-axis motor current. A brief 30~Nm step test confirmed the maximum torque (Fig.~\ref{fig:constant_torque_speed}.b).

\subsection{System Identification}

We used a least squares regression problem to predict the torque sensor values resulting from the constant torque and speed test. We found a model of the form,
\begin{align}
\tau_{\mathrm{output}} = & b+g_r(k_{\tau}- k_{n}\left \| I_{q} \right \|)I_{q}\nonumber\\&- (f_{C}+f_{g} \mathrm{abs}(I_{q})) \mathrm{sign}(\dot{\theta}), \label{sysIDeq}
\end{align}
with $\tau_{\mathrm{output}}$ the measured output torque, $b$ a torque sensor bias, $k_\tau$ the nominal torque constant, $k_n$ a nonlinear current to torque relationship parameter, $f_C$ a Coulomb friction term, $f_g$ a gear friction term, and $I_q$ the q-axis current. We filtered both the regressor matrix and the target torque with a second order low pass filter (2 Hz, $\zeta=.7$). This 2 Hz frequency was chosen to focus the model on accurately predicting steady-state torque rather than the torque and current transients.

We found a model with 95\% of errors less than 0.39 Nm. This model's nonlinear torque constant $k_\tau + k_n\|I_q\|$ varied linearly between 0.147 $\mathrm{Nm/A}$ near zero current to 0.125 $\mathrm{Nm/A}$ at 20 Amps (q-axis). Coulomb and gear friction were modeled at 0.37 Nm and  8.8\% of the nominal torque, respectively. This implies a frictional backdriving resistance at high torque values above what we were able to measure in the static backdriving test. 

\subsection{Identifying Inertia}
The constant torque and speed test is not well conditioned to estimate the reflected inertia, however the dynamic backdrive test shows the influence of reflected inertia. We performed a second regression on the first model's residual when applied to predicting the result of the backdrive test. The filter from the previous identification was re-used. 

The second model included only a bias and a reflected inertia term. The second model regression reduced the RMSE in the dynamic backdrive trial from 0.28~Nm to 0.061~Nm. For reference, the RMSE of the filtered torque signal itself in the dynamic backdrive trial is 0.36~Nm, which is to say that the dynamic backdrive trial is dominated by the effect of inertia which the first model fails to describe. The reflected inertia estimate was $92.11\ \mathrm{kg\cdot cm^2}$, less than half of the reflected inertia of our previous knee exoskeleton actuator design ($200.9\ \mathrm{kg\cdot cm^2}$) \cite{Zhu2021}.

% \subsection{Benchtop Testing Results}

% Benchtop experiments demonstrated a static backdrive torque of under 0.5~Nm using a manual torque wrench, and a peak backdrive torque of ===RESULTHERE~Nm during the dynamic manual backdrive test using the FUTEK load cell. The velocity and torque trajectories from this test are presented in Fig.~\ref{fig:dynBD}.

% The system identification test produced a model for open-loop torque control of the form 
% \begin{multline}
% \tau_{\mathrm{output}}=b+(k_{\tau}- k_{n}\left \| I_{q} \right \|)I_{q}-\\(f_{C}+f_{g} \mathrm{abs}(I_{q})) \mathrm{sign}(\dot{\theta}) \label{sysIDeq}
% \end{multline} 
% taking into account q-axis current ($I_{q}$), nominal q-axis actuator torque constant ($k_{\tau}$=1.32),  non-linearity ($k_{n}$=0.0106), Coulomb friction ($f_{c}$=0.37), gear friction ($f_{g}$=0.111) proportional to $I_{q}$, and the direction of velocity ($\dot{\theta}$). This model predicts output torque to within 0.25~Nm at frequencies below 2~Hz (second order filter, $\zeta=0.7$). 

%\begin{figure}
%\centering
%\includegraphics[width = 0.4\textwidth]{FrictionCharacterizationPlaceholder.png}
%\vspace{-1mm}
%\caption{Linear regression model of output torque as a function of independent variables velocity sign and input current (q-axis).} \label{fig:fric}
%\vspace{-2mm}
%\end{figure}

\section{Human Subject Experiment} \label{Sec:ExpMethods}

\subsection{Methods} 

The experimental protocol was approved by the Institutional Review Board of the University of Michigan (HUM00164931). A single able-bodied human subject (male, 34 years old, 82~kg, 178~cm tall) was recruited for the study. The $l_h$ and $l_t$ values specified above were left unchanged, though the subject's anatomical thigh length is shorter. This effectively increases the assistance ratio by a small factor. Four experimental modes were tested: bare mode (without exoskeleton) and hip-only, knee-only, and hip-knee exoskeleton configurations (see Fig. \ref{fig:configs}). The modular exoskeleton counteracted 20\% of the gravitational torques on the supported joints (hip and/or knee). Per experimental mode, the subject performed 20 alternating lifting and lowering (L\&L) squats using a 12.5~kg mass, which was approximately 15\% of the subject's body mass. A 60~beats/min metronome was used to guide the subject's squatting pace in all trials. The L\&L sequence started with the subject standing upright on the first beat. On the second beat, the subject achieved a squat position, and either grasped or released the weight for a lifting or lowering trial respectively. The sequence ended with the subject returning to an upright standing posture on the third beat. Approximately 2-3 seconds (2-3 beats) of rest was taken after the third beat to prevent fatigue. Further, about 10-15 minutes of rest was provided between sets (experimental modes). The trials were performed over floor-embedded force plates (AMTI, Massachusetts, USA). The subject stood with each foot on a separate force plate, and was provided visual feedback of the vertical ground reaction forces. Visual feedback was provided to ensure the subject retained symmetrical loading of their legs and prevent them from favoring either the assisted or unassisted limb.

After appropriate skin preparation, six electromyography (EMG) electrodes (Trigno Avanti, Delsys, Massachusetts, USA) were taped onto the subject's right limb over the vastus medials oblique (VM), vastus lateralis (VL), rectus femoris (RF), biceps femoris (BF), semitendinosus (ST), and gluteus maximus (GM) to assess the effect of the exoskeleton assistance on muscle activation.  A maximum voluntary contraction (MVC) procedure which consisted of explosive squats (dynamic contraction) as well as maximal isometric contraction against manual resistance was completed and used to normalize EMG data to \%MVC. L\&L trials were cropped into individual repetitions by utilizing the thigh orientation angle (acquired from additional sensors built into the RF EMG sensor). Each muscle's EMG was demeaned, bandpass filtered (20 - 200~Hz), and smoothed with a moving 100~ms window RMS filter. After normalizing the EMG to \%MVC, the integral with respect to time was calculated to represent muscular effort as \%MVC$\cdot$s for each repetition. Further, the peak muscle activation during each repetition was found.  

\subsection{Results} 
\begin{figure}[htb]
\centering
\includegraphics[]{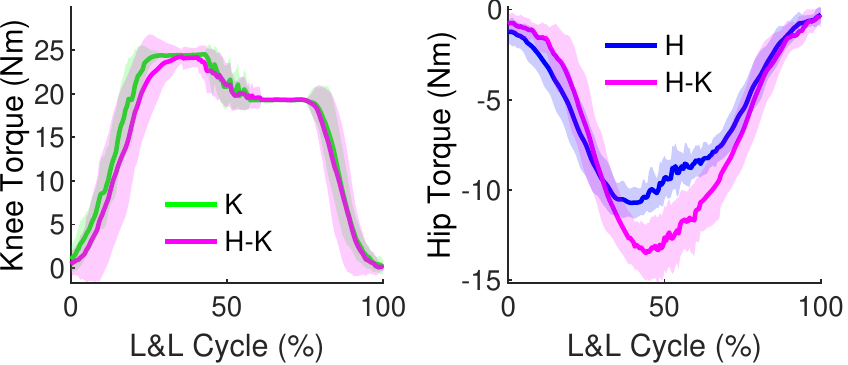}
\caption{Knee (left) and hip (right) torques for knee-only~(K), hip-only~(H), and hip-knee~(H-K) exoskeleton configurations during L\&L tasks.} \label{fig:LL_torques}
\vspace{-2mm}
\end{figure}

\begin{figure}[htb]
\centering
\includegraphics[]{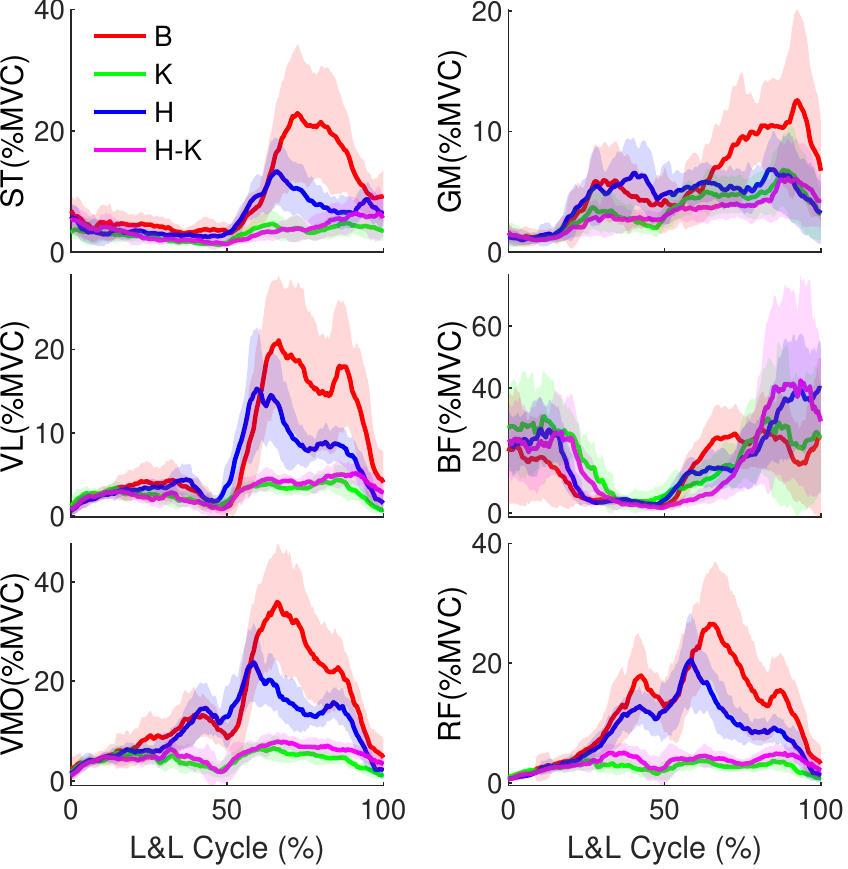}
\caption{Ensemble average muscle activation profiles for bare~(B) vs. exoskeleton configurations: knee-only~(K), hip-only~(H), and hip-knee~(HK) exoskeleton configurations during L\&L task.} \label{fig:LL_emg}
\vspace{-2mm}
\end{figure}

The highly backdrivable M-BLUE system was able to deliver meaningful assistance torques to the knee and/or hip during the L\&L task, which resulted in considerable reductions in activation of the squat musculature. Torques up to 26 Nm were provided by the knee actuator, while torques up to 15 Nm were provided by the hip actuator (see Fig. \ref{fig:LL_torques}). Considering the ensemble average EMG plots (Fig.~\ref{fig:LL_emg}), it can be seen that the assistance torques were generally synchronous with the knee and hip extensor muscle activations, especially during the concentric part of the motion (50-100\% L\&L cycle). Muscle activation was generally quite low during the eccentric part of the motion (0-50\% L\&L cycle) for all modes. Importantly, muscle activations (barring the BF) in the concentric half of the LL cycle were substantially lower for all exoskeleton configurations when compared to bare mode. The greatest reductions in muscle activations were apparent with configurations that included the knee module, \textit{i.e.}, knee only and hip-knee configurations. Moreover, addition of the hip module to the knee module did not result in further EMG reductions. Yet, the hip only configuration facilitates meaningful EMG reductions compared to bare. Ultimately, the summary metrics (muscle effort and peak activation) show M-BLUE was beneficial in all three configurations. With the exception of the BF, the mean effort and mean peak EMG of all muscles tested were substantially lower with exoskeleton assistance than without it (Tabs. \ref{EMGcompstabeffort}-\ref{EMGcompstabpeak}). The BF activations were quite similar for all modes, with the standard deviations overlapping throughout the L\&L cycle.

%\begin{figure}[h]
%\centering
%\includegraphics[width = 0.48\textwidth,trim={0 0.3mm 0 0},clip]{AB01EMGLL_barMVC.pdf}
%\caption{Muscle effort and peak activation comparisons between bare~(B), knee-only~(K), hip-only~(H), and hip-knee~(H-K) exoskeleton configurations during L\&L tasks.} \label{fig:LL_bar}
%\vspace{-2mm}
%\end{figure}

\begin{table}[!htb]
\caption{Muscle effort comparisons for bare vs exoskeleton configurations: knee only, hip only, and hip-knee showing mean effort ($\pm$~SD).}\vspace{-2mm}
\label{EMGcompstabeffort}
\begin{tabular}{l|llll}
   & Bare    & Knee    & Hip     & Hip-Knee \\ \hline
VM & 33.2 (6.6) & 10.5 (0.9) & 27.7 (2.9) & 13.4 (2.2)  \\
RF & 26.2 (3.8) & 6.5 (0.9)  & 21.2 (2.6) & 9.0 (1.8)   \\
VL & 18.0 (3.9) & 6.7 (0.6)  & 14.7 (1.9) & 7.9 (0.8)   \\
BF & 31.5 (6.4) & 41.8 (7.0) & 42.1 (7.2) & 42.9 (11.3)  \\
ST & 19.9 (3.4) & 7.4 (0.9)  & 14.3 (2.1) & 8.8 (1.5)   \\
GM & 12.6 (1.8) & 8.5 (1.3)  & 11.4 (1.0) & 8.2 (1.2)  
\end{tabular}
\end{table}

\begin{table}[!htb]
\caption{Peak muscle activation comparisons for bare vs exoskeleton configurations: knee only, hip only, and hip-knee showing mean peak ($\pm$~SD).}\vspace{-2mm}
\label{EMGcompstabpeak}
\begin{tabular}{l|llll}
   & Bare    & Knee    & Hip     & Hip-Knee \\ \hline
VM & 44.1 (10.8) & 9.3 (1.5)  & 29.4 (6.6) & 10.4 (1.9)  \\
RF & 33.8 (8.3) & 6.1 (1.3)  & 24.4 (6.4) & 7.4 (2.1)   \\
VL & 29.3 (8.2) & 6.0 (1.1)  & 19.8 (6.9) & 6.8 (1.2)   \\
BF & 50.4 (14.1) & 58.1 (11.6) & 53.6 (13.0) & 66.5 (23.9)  \\
ST & 29.6 (9.7) & 7.7 (1.9)  & 16.0 (5.1) & 10.5 (3.4)  \\
GM & 17.0 (5.6) & 9.1 (3.2)  & 11.7 (2.8) & 8.5 (2.9)  
\end{tabular}
\end{table}

\section{Discussion} \label{Sec:Discussion}
The presented design and performance characteristics of M-BLUE demonstrates its potential impact for clinical applications. The benchtop and human subject experiments indicate this design paradigm is appropriate for assisting the musculature of individuals with mild to moderate impairments. These aspects of the study are discussed below, and serve to motivate future work with M-BLUE.

The presented results motivate future work related to the physical design of M-BLUE. The molded plastic and sheet metal structure (both the stock components and the custom revisions) adequately transferred the loads to which they were subjected, and may be capable of handling even higher output torques. Future experimentation will confirm whether the current design paradigm is sufficient for the upper limit of the actuators' torque capabilities. The rapid don/doff time and ease of adjustment provided by the commercial orthotic interface will facilitate optimally productive research sessions, particularly in instances where the wearer's time availability is a limiting factor. Some design refinements may further improve the physical system through the introduction of mechanisms to prevent distal sliding of the knee brace and to allow passive ab/adduction of the hip joint to occur more freely. Considerations will also be made for interfacing M-BLUE with the commercially-available Dephy ExoBoot \cite{Ingraham2020,Mooney2014} in cases where plantarflexion assistance is merited.

The benchtop assessment of the M-BLUE actuators indicated that these actuators are well-suited to the application of clinically-viable, partial-assist, lower-limb exoskeletons in terms of cost, comfort, and ease of control. The low backdrive torque exhibited during zero current control, even during highly dynamic perturbations, suggests that M-BLUE will be capable of assisting users across a variety of tasks (overground walking, stairs, etc.), not only those (such as L\&L) which occur on a more gradual timescale. The precision with which the derived torque output model predicts the actuator's behavior across torques and speeds indicates that M-BLUE is capable of executing open-loop control reliably. This has significant implications on cost and weight, as it eliminates the need for additional sensor hardware to close the feedback loop and achieve high-performance control.

The human subject experiment yielded promising results, with the majority of the investigated muscles showing a clear reduction of both peak muscle activation and effort, particularly in configurations that include the knee module. The knee actuator alone produced substantial EMG reduction, with prominent decreases visible in most of the muscles' activation profiles, while the addition of the hip actuator did not substantively improve performance. This can likely be attributed to the greater knee joint deflection required to perform a squat lift (relative to the hip joint). Based on these findings, it is probable that the knee-only configuration has the greatest potential for reducing fatigue in repetitive squat lifting tasks. It is possible that performing activities that demand more work of the hips, such as a stoop lift \cite{Hsiang1997}, would produce more impressive results for the hip module. Alternatively, a higher assistance ratio for the hip controller could improve the module's efficacy by increasing the torque output for the same motions and bring the torque output at the hip in line with what is seen at the knee.

% The reduction in peak muscular activation motivates future applications to knee osteoarthritis (OA) \cite{MedranoRouseThomas2021TMRB}, where pain in the patellofemoral joint is related to quadriceps force. By reducing peak muscle activation/force, patients may be able to perform demanding tasks such as sit-stand and stair climbing with less pain. Importantly, this encourages a more active lifestyle, which is recommended for controlling the progression of knee OA. On the other hand, the demonstrated reduction in muscle effort is beneficial in the context of movement augmentation. For example, assisting the quadriceps may encourage and prolong the use of the recommended squat technique in workplace L\&L tasks known to cause lower back pain \cite{Hsiang1997}.

This modular orthotic technology has the potential for clinical impact in several different populations with mild to moderate joint-specific impairments.
First, M-BLUE could provide a novel non-surgical treatment for osteoarthritic pain associated with compressive loads at the hip or knee. We have previously found that lightweight exoskeletons without a parallel structure are well suited for reducing compressive joint loads due to muscle contraction \cite{MedranoRouseThomas2021TMRB}. Conventional OA braces unload the medial or lateral compartments of the knee but cannot unload the patellofemoral compartment, which can only be done by reducing quadriceps muscle force. The hip orthosis will make orthotic treatments a viable option for hip OA, which currently lacks such treatments. By reducing peak muscle activation/force, patients may be able to perform demanding tasks such as sit-stand and stair climbing with less pain. Importantly, this encourages a more active lifestyle, which is recommended for controlling the progression of knee OA.
Second, the validation study with L\&L suggests that the presented system can reduce muscular fatigue leading to risky lifting posture \cite{trafimow1993effects}. Assisting the quadriceps in particular may encourage and prolong the use of the recommended squat technique in workplace L\&L tasks known to cause lower back pain \cite{Hsiang1997}. Thus, bilateral configurations of M-BLUE could serve to prevent chronic overuse musculoskeletal injuries in the workplace, which cost employers billions every year \cite{CDC:WorkMSD}. 
Third, targeting specific stroke deficits may reduce compensations at more functional joints, reducing the associated cognitive burden, energy expenditure, fall risk, and joint overuse \cite{Woolley2001}. This motivates applications in mobile gait training to increase flexibility in the clinic and dosage through community-based therapy.
And fourth, bilateral hip and knee orthoses could improve mobility and quality of life for individuals with age-related immobility \cite{grimmer2019mobility}. Some of these populations would further benefit from integration of the Dephy ExoBoot \cite{Ingraham2020,Mooney2014} as an ankle module for plantarflexion assistance. These investigations are left to future work.

\section{Conclusion}

This paper introduced the design and validation of a modular exoskeleton system (M-BLUE) that leverages highly backdrivable, torque-dense actuators and commercial orthoses. The benchtop tests verified the low mechanical impedance of the actuator and the accuracy of open-loop current control as a means to achieve desired torques via a regression-based model. The human subject experiment confirmed the compatibility of our modular exoskeleton system with an able-bodied wearer, and demonstrated the efficacy of the device at substantially reducing muscle activation during a demanding L\&L task.

Future work will include the extension of the modular design to encompass bilateral and ankle configurations, as well as implementation of more advanced controllers. Future human subject experiments will investigate the performance of these configurations and controllers on larger populations of able-bodied subjects, along with pathological populations such as osteoarthritis and age-related frailty. We will also compile and disseminate instructive guides on the disassembly of stock brace configurations and installation of M-BLUE hardware to assist clinicians in the implementation of modular exoskeletons.

% Use future work to admit that we did not do something?
% Potential future work topics (specific mention of any of these?): design refinement to increase ease of ambulation through passive hip ab/adduction; knee only configuration updates for suspension and migration management (calf belly strap, constant force spring connecting to waist, passive AFO attachment to ground the device to the foot and prevent distal slipping); higher torques and thermal analysis; 
\bibliographystyle{IEEEtran}
\bibliography{refs.bib}

\end{document}

%% file: M-BLUE-ASSY-MedResImg.pdf_tex
%% Accompanies image file 'M-BLUE-ASSY-MedResImg.pdf' (pdf, eps, ps)
\begingroup%
  \makeatletter%
  \providecommand\color[2][]{%
    \errmessage{(Inkscape) Color is used for the text in Inkscape, but the package 'color.sty' is not loaded}%
    \renewcommand\color[2][]{}%
  }%
  \providecommand\transparent[1]{%
    \errmessage{(Inkscape) Transparency is used (non-zero) for the text in Inkscape, but the package 'transparent.sty' is not loaded}%
    \renewcommand\transparent[1]{}%
  }%
  \providecommand\rotatebox[2]{#2}%
  \newcommand*\fsize{\dimexpr\f@size pt\relax}%
  \newcommand*\lineheight[1]{\fontsize{\fsize}{#1\fsize}\selectfont}%
  \ifx\svgwidth\undefined%
    \setlength{\unitlength}{468bp}%
    \ifx\svgscale\undefined%
      \relax%
    \else%
      \setlength{\unitlength}{\unitlength * \real{\svgscale}}%
    \fi%
  \else%
    \setlength{\unitlength}{\svgwidth}%
  \fi%
  \global\let\svgwidth\undefined%
  \global\let\svgscale\undefined%
  \makeatother%
  \begin{picture}(1,0.64615382)%
    \lineheight{1}%
    \setlength\tabcolsep{0pt}%
    \put(0,0){\includegraphics[width=\unitlength,page=1]{M-BLUE-ASSY-MedResImg.pdf}}%
    \put(0.10448295,0.56980105){\color[rgb]{0,0,0}\makebox(0,0)[lt]{\lineheight{1.25}\smash{\begin{tabular}[t]{l}a1\end{tabular}}}}%
    \put(0.24564187,0.61297455){\color[rgb]{0,0,0}\makebox(0,0)[lt]{\lineheight{1.25}\smash{\begin{tabular}[t]{l}a2\end{tabular}}}}%
    \put(0.38680078,0.61297455){\color[rgb]{0,0,0}\makebox(0,0)[lt]{\lineheight{1.25}\smash{\begin{tabular}[t]{l}c1\end{tabular}}}}%
    \put(0.52795965,0.56980105){\color[rgb]{0,0,0}\makebox(0,0)[lt]{\lineheight{1.25}\smash{\begin{tabular}[t]{l}c2\end{tabular}}}}%
    \put(0.6691186,0.61297455){\color[rgb]{0,0,0}\makebox(0,0)[lt]{\lineheight{1.25}\smash{\begin{tabular}[t]{l}e1\end{tabular}}}}%
    \put(0.81027747,0.49847069){\color[rgb]{0,0,0}\makebox(0,0)[lt]{\lineheight{1.25}\smash{\begin{tabular}[t]{l}e2\end{tabular}}}}%
    \put(0.10448295,0.30043527){\color[rgb]{0,0,0}\makebox(0,0)[lt]{\lineheight{1.25}\smash{\begin{tabular}[t]{l}b1\end{tabular}}}}%
    \put(0.24564187,0.31545229){\color[rgb]{0,0,0}\makebox(0,0)[lt]{\lineheight{1.25}\smash{\begin{tabular}[t]{l}b2\end{tabular}}}}%
    \put(0.38680078,0.31545229){\color[rgb]{0,0,0}\makebox(0,0)[lt]{\lineheight{1.25}\smash{\begin{tabular}[t]{l}d1\end{tabular}}}}%
    \put(0.52795965,0.30043527){\color[rgb]{0,0,0}\makebox(0,0)[lt]{\lineheight{1.25}\smash{\begin{tabular}[t]{l}d2\end{tabular}}}}%
    \put(0.08060359,0.32737004){\color[rgb]{0.90588235,0.51372549,0.38039216}\rotatebox{90}{\makebox(0,0)[t]{\lineheight{1.25}\smash{\begin{tabular}[t]{c}original uprights\end{tabular}}}}}%
    \put(0,0){\includegraphics[width=\unitlength,page=2]{M-BLUE-ASSY-MedResImg.pdf}}%
    \put(0.36047995,0.32325395){\color[rgb]{0.90588235,0.51372549,0.38039216}\rotatebox{90}{\makebox(0,0)[t]{\lineheight{1.25}\smash{\begin{tabular}[t]{c}replacement uprights\end{tabular}}}}}%
    \put(0,0){\includegraphics[width=\unitlength,page=3]{M-BLUE-ASSY-MedResImg.pdf}}%
    \put(0.78378682,0.3505406){\color[rgb]{0.90588235,0.51372549,0.38039216}\rotatebox{87.234966}{\makebox(0,0)[t]{\lineheight{1.25}\smash{\begin{tabular}[t]{c}adjustable-length thigh upright\end{tabular}}}}}%
    \put(0,0){\includegraphics[width=\unitlength,page=4]{M-BLUE-ASSY-MedResImg.pdf}}%
  \end{picture}%
\endgroup%

%% file: DynamicBackdriveTwoAxesInOne.pdf_tex
%% Accompanies image file 'DynamicBackdriveTwoAxesInOne.pdf' (pdf, eps, ps)
\begingroup%
  \makeatletter%
  \providecommand\color[2][]{%
    \errmessage{(Inkscape) Color is used for the text in Inkscape, but the package 'color.sty' is not loaded}%
    \renewcommand\color[2][]{}%
  }%
  \providecommand\transparent[1]{%
    \errmessage{(Inkscape) Transparency is used (non-zero) for the text in Inkscape, but the package 'transparent.sty' is not loaded}%
    \renewcommand\transparent[1]{}%
  }%
  \providecommand\rotatebox[2]{#2}%
  \newcommand*\fsize{\dimexpr\f@size pt\relax}%
  \newcommand*\lineheight[1]{\fontsize{\fsize}{#1\fsize}\selectfont}%
  \ifx\svgwidth\undefined%
    \setlength{\unitlength}{252bp}%
    \ifx\svgscale\undefined%
      \relax%
    \else%
      \setlength{\unitlength}{\unitlength * \real{\svgscale}}%
    \fi%
  \else%
    \setlength{\unitlength}{\svgwidth}%
  \fi%
  \global\let\svgwidth\undefined%
  \global\let\svgscale\undefined%
  \makeatother%
  \begin{picture}(1,0.2)%
    \lineheight{1}%
    \setlength\tabcolsep{0pt}%
    \put(0,0){\includegraphics[width=\unitlength,page=1]{DynamicBackdriveTwoAxesInOne.pdf}}%
    \put(0.23474322,0.03139362){\color[rgb]{0,0,0}\makebox(0,0)[lt]{\lineheight{1.25}\smash{\begin{tabular}[t]{l}5\end{tabular}}}}%
    \put(0,0){\includegraphics[width=\unitlength,page=2]{DynamicBackdriveTwoAxesInOne.pdf}}%
    \put(0.36692352,0.03139362){\color[rgb]{0,0,0}\makebox(0,0)[lt]{\lineheight{1.25}\smash{\begin{tabular}[t]{l}10\end{tabular}}}}%
    \put(0,0){\includegraphics[width=\unitlength,page=3]{DynamicBackdriveTwoAxesInOne.pdf}}%
    \put(0.50371105,0.03139362){\color[rgb]{0,0,0}\makebox(0,0)[lt]{\lineheight{1.25}\smash{\begin{tabular}[t]{l}15\end{tabular}}}}%
    \put(0,0){\includegraphics[width=\unitlength,page=4]{DynamicBackdriveTwoAxesInOne.pdf}}%
    \put(0.64049855,0.03139362){\color[rgb]{0,0,0}\makebox(0,0)[lt]{\lineheight{1.25}\smash{\begin{tabular}[t]{l}20\end{tabular}}}}%
    \put(0,0){\includegraphics[width=\unitlength,page=5]{DynamicBackdriveTwoAxesInOne.pdf}}%
    \put(0.77728609,0.03139362){\color[rgb]{0,0,0}\makebox(0,0)[lt]{\lineheight{1.25}\smash{\begin{tabular}[t]{l}25\end{tabular}}}}%
    \put(0.5136359,0.00273144){\color[rgb]{0,0,0}\makebox(0,0)[t]{\lineheight{1.25}\smash{\begin{tabular}[t]{c}Time (s)\end{tabular}}}}%
    \put(0,0){\includegraphics[width=\unitlength,page=6]{DynamicBackdriveTwoAxesInOne.pdf}}%
    \put(0.07253637,0.06895726){\color[rgb]{0,0,0}\makebox(0,0)[lt]{\lineheight{1.25}\smash{\begin{tabular}[t]{l}2\end{tabular}}}}%
    \put(0,0){\includegraphics[width=\unitlength,page=7]{DynamicBackdriveTwoAxesInOne.pdf}}%
    \put(0.07253637,0.10020092){\color[rgb]{0,0,0}\makebox(0,0)[lt]{\lineheight{1.25}\smash{\begin{tabular}[t]{l}1\end{tabular}}}}%
    \put(0,0){\includegraphics[width=\unitlength,page=8]{DynamicBackdriveTwoAxesInOne.pdf}}%
    \put(0.07253071,0.13144458){\color[rgb]{0,0,0}\makebox(0,0)[lt]{\lineheight{1.25}\smash{\begin{tabular}[t]{l}0\end{tabular}}}}%
    \put(0,0){\includegraphics[width=\unitlength,page=9]{DynamicBackdriveTwoAxesInOne.pdf}}%
    \put(0.07253071,0.16268825){\color[rgb]{0,0,0}\makebox(0,0)[lt]{\lineheight{1.25}\smash{\begin{tabular}[t]{l}1\end{tabular}}}}%
    \put(0.02623237,0.10835951){\color[rgb]{0,0,0}\rotatebox{90}{\makebox(0,0)[t]{\lineheight{1.25}\smash{\begin{tabular}[t]{c}Torque (Nm)\end{tabular}}}}}%
    \put(0,0){\includegraphics[width=\unitlength,page=10]{DynamicBackdriveTwoAxesInOne.pdf}}%
    \put(0.92111222,0.06750454){\color[rgb]{0.12156863,0.46666667,0.70588235}\makebox(0,0)[lt]{\lineheight{1.25}\smash{\begin{tabular}[t]{l}50\end{tabular}}}}%
    \put(0,0){\includegraphics[width=\unitlength,page=11]{DynamicBackdriveTwoAxesInOne.pdf}}%
    \put(0.90897161,0.12348475){\color[rgb]{0.12156863,0.46666667,0.70588235}\makebox(0,0)[lt]{\lineheight{1.25}\smash{\begin{tabular}[t]{l}0\end{tabular}}}}%
    \put(0,0){\includegraphics[width=\unitlength,page=12]{DynamicBackdriveTwoAxesInOne.pdf}}%
    \put(0.90897161,0.17946496){\color[rgb]{0.12156863,0.46666667,0.70588235}\makebox(0,0)[lt]{\lineheight{1.25}\smash{\begin{tabular}[t]{l}50\end{tabular}}}}%
    \put(0.98815988,0.11085824){\color[rgb]{0.12156863,0.46666667,0.70588235}\rotatebox{90}{\makebox(0,0)[t]{\lineheight{1.25}\smash{\begin{tabular}[t]{c}Position (deg)\end{tabular}}}}}%
    \put(0,0){\includegraphics[width=\unitlength,page=13]{DynamicBackdriveTwoAxesInOne.pdf}}%
  \end{picture}%
\endgroup%

%% file: torque_speed_trial.pdf_tex
%% Accompanies image file 'torque_speed_trial.pdf' (pdf, eps, ps)
\begingroup%
  \makeatletter%
  \providecommand\color[2][]{%
    \errmessage{(Inkscape) Color is used for the text in Inkscape, but the package 'color.sty' is not loaded}%
    \renewcommand\color[2][]{}%
  }%
  \providecommand\transparent[1]{%
    \errmessage{(Inkscape) Transparency is used (non-zero) for the text in Inkscape, but the package 'transparent.sty' is not loaded}%
    \renewcommand\transparent[1]{}%
  }%
  \providecommand\rotatebox[2]{#2}%
  \newcommand*\fsize{\dimexpr\f@size pt\relax}%
  \newcommand*\lineheight[1]{\fontsize{\fsize}{#1\fsize}\selectfont}%
  \ifx\svgwidth\undefined%
    \setlength{\unitlength}{252bp}%
    \ifx\svgscale\undefined%
      \relax%
    \else%
      \setlength{\unitlength}{\unitlength * \real{\svgscale}}%
    \fi%
  \else%
    \setlength{\unitlength}{\svgwidth}%
  \fi%
  \global\let\svgwidth\undefined%
  \global\let\svgscale\undefined%
  \makeatother%
  \begin{picture}(1,0.5)%
    \lineheight{1}%
    \setlength\tabcolsep{0pt}%
    \put(0,0){\includegraphics[width=\unitlength,page=1]{torque_speed_trial.pdf}}%
    \put(0.0490934,0.40849147){\color[rgb]{0,0,0}\rotatebox{90}{\makebox(0,0)[t]{\lineheight{1.25}\smash{\begin{tabular}[t]{c}Torque (Nm)\end{tabular}}}}}%
    \put(0.41161701,0.00695525){\color[rgb]{0,0,0}\makebox(0,0)[t]{\lineheight{1.25}\smash{\begin{tabular}[t]{c}Time (s)\end{tabular}}}}%
    \put(0.0490934,0.17819461){\color[rgb]{0,0,0}\rotatebox{90}{\makebox(0,0)[t]{\lineheight{1.25}\smash{\begin{tabular}[t]{c}Speed (deg/s)\end{tabular}}}}}%
    \put(0.11397782,0.33720168){\color[rgb]{0,0,0}\makebox(0,0)[rt]{\lineheight{1.25}\smash{\begin{tabular}[t]{r}-25\end{tabular}}}}%
    \put(0.11397782,0.10867811){\color[rgb]{0,0,0}\makebox(0,0)[rt]{\lineheight{1.25}\smash{\begin{tabular}[t]{r}-250\end{tabular}}}}%
    \put(0.11397782,0.23021114){\color[rgb]{0,0,0}\makebox(0,0)[rt]{\lineheight{1.25}\smash{\begin{tabular}[t]{r}250\end{tabular}}}}%
    \put(0.11397782,0.1694446){\color[rgb]{0,0,0}\makebox(0,0)[rt]{\lineheight{1.25}\smash{\begin{tabular}[t]{r}0\end{tabular}}}}%
    \put(0.11397782,0.39433258){\color[rgb]{0,0,0}\makebox(0,0)[rt]{\lineheight{1.25}\smash{\begin{tabular}[t]{r}0\end{tabular}}}}%
    \put(0.11397782,0.45146349){\color[rgb]{0,0,0}\makebox(0,0)[rt]{\lineheight{1.25}\smash{\begin{tabular}[t]{r}25\end{tabular}}}}%
    \put(0.15690853,0.03975511){\color[rgb]{0,0,0}\makebox(0,0)[t]{\lineheight{1.25}\smash{\begin{tabular}[t]{c}0\end{tabular}}}}%
    \put(0.24290008,0.03975511){\color[rgb]{0,0,0}\makebox(0,0)[t]{\lineheight{1.25}\smash{\begin{tabular}[t]{c}25\end{tabular}}}}%
    \put(0.32889162,0.03975511){\color[rgb]{0,0,0}\makebox(0,0)[t]{\lineheight{1.25}\smash{\begin{tabular}[t]{c}50\end{tabular}}}}%
    \put(0.41488316,0.03975511){\color[rgb]{0,0,0}\makebox(0,0)[t]{\lineheight{1.25}\smash{\begin{tabular}[t]{c}75\end{tabular}}}}%
    \put(0.50087479,0.03975511){\color[rgb]{0,0,0}\makebox(0,0)[t]{\lineheight{1.25}\smash{\begin{tabular}[t]{c}100\end{tabular}}}}%
    \put(0.58686629,0.03975511){\color[rgb]{0,0,0}\makebox(0,0)[t]{\lineheight{1.25}\smash{\begin{tabular}[t]{c}125\end{tabular}}}}%
    \put(0.67285787,0.03975511){\color[rgb]{0,0,0}\makebox(0,0)[t]{\lineheight{1.25}\smash{\begin{tabular}[t]{c}150\end{tabular}}}}%
    \put(0.20040887,0.246831){\color[rgb]{0,0,0}\makebox(0,0)[lt]{\lineheight{1.25}\smash{\begin{tabular}[t]{l}desired\end{tabular}}}}%
    \put(0.20040887,0.21670747){\color[rgb]{0,0,0}\makebox(0,0)[lt]{\lineheight{1.25}\smash{\begin{tabular}[t]{l}measured\end{tabular}}}}%
    \put(0.8470063,0.01520844){\color[rgb]{0,0,0}\makebox(0,0)[lt]{\lineheight{1.25}\smash{\begin{tabular}[t]{l}Time (s)\end{tabular}}}}%
    \put(0.76281702,0.09772874){\color[rgb]{0,0,0}\makebox(0,0)[rt]{\lineheight{1.25}\smash{\begin{tabular}[t]{r}0\end{tabular}}}}%
    \put(0.76111771,0.18180217){\color[rgb]{0,0,0}\makebox(0,0)[rt]{\lineheight{1.25}\smash{\begin{tabular}[t]{r}10\end{tabular}}}}%
    \put(0.76252783,0.26587556){\color[rgb]{0,0,0}\makebox(0,0)[rt]{\lineheight{1.25}\smash{\begin{tabular}[t]{r}20\end{tabular}}}}%
    \put(0.76245553,0.34994894){\color[rgb]{0,0,0}\makebox(0,0)[rt]{\lineheight{1.25}\smash{\begin{tabular}[t]{r}30\end{tabular}}}}%
    \put(0.70339852,0.24930601){\color[rgb]{0,0,0}\rotatebox{90}{\makebox(0,0)[t]{\lineheight{1.25}\smash{\begin{tabular}[t]{c}Torque (Nm)\end{tabular}}}}}%
    \put(0.81805138,0.45910519){\color[rgb]{0,0,0}\makebox(0,0)[lt]{\lineheight{1.25}\smash{\begin{tabular}[t]{l}Torque (Futek)\end{tabular}}}}%
    \put(0.81205341,0.42289036){\color[rgb]{0,0,0}\makebox(0,0)[lt]{\lineheight{1.25}\smash{\begin{tabular}[t]{l}Desired torque\end{tabular}}}}%
    \put(0.8459925,0.04403647){\color[rgb]{0,0,0}\makebox(0,0)[t]{\lineheight{1.25}\smash{\begin{tabular}[t]{c}0\end{tabular}}}}%
    \put(0.93857157,0.04403647){\color[rgb]{0,0,0}\makebox(0,0)[t]{\lineheight{1.25}\smash{\begin{tabular}[t]{c}0.1\end{tabular}}}}%
    \put(0.00475904,0.46980573){\color[rgb]{0,0,0}\makebox(0,0)[lt]{\lineheight{1.25}\smash{\begin{tabular}[t]{l}a.\end{tabular}}}}%
    \put(0.70563534,0.46980573){\color[rgb]{0,0,0}\makebox(0,0)[rt]{\lineheight{1.25}\smash{\begin{tabular}[t]{r}b.\end{tabular}}}}%
  \end{picture}%
\endgroup%